\documentclass{article}
\usepackage{spconf,amsmath,graphicx}
\usepackage{epsfig}
\usepackage{amssymb}
\usepackage{array}
\usepackage{float}
\usepackage{color}
\usepackage{multirow}
\usepackage{url}
\usepackage[normalem]{ulem}
\usepackage{microtype}

\usepackage{mathtools}
\usepackage{rotating}
\DeclarePairedDelimiter\norm{\lVert}{\rVert}
\DeclareMathOperator*{\argmax}{argmax} 
\DeclareMathOperator*{\argmin}{argmin}
\newcommand{\net}{\mathcal{N}}
\DeclareRobustCommand\onedot{\futurelet\@let@token\@onedot}
\def\@onedot{\ifx\@let@token.\else.\null\fi\xspace}
 
\usepackage{bibspacing}
\def\x{{\mathbf x}}

\newcommand{\comment}[1]{}

\name{Kanchana Vaishnavi Gandikota \qquad  Paramanand Chandramouli \qquad Michael Moeller}
\address{Department of Computer Science, University of Siegen\\
{\tt\small \{kanchana.gandikota,  paramanand.chandramouli, michael.moeller\}@uni-siegen.de}}
\title{On  Adversarial Robustness of Deep Image Deblurring}
\begin{document}
%
\maketitle
\begin{abstract}
Recent approaches employ deep learning-based solutions for the recovery of a sharp image from its blurry observation. This paper introduces adversarial attacks against deep learning-based image deblurring methods and evaluates the robustness of these neural networks to untargeted and targeted  attacks. We demonstrate that imperceptible distortion can significantly degrade the performance of state-of-the-art deblurring networks, even producing  drastically different content in the output, indicating the strong need to include adversarially robust training not only in classification but also for image recovery.  
\end{abstract}
\begin{keywords}
adversarial attack, image deblurring
\end{keywords}
\section{Introduction}
\label{sec:intro}
Image deblurring which aims at recovering sharp images from blurred inputs is an important and well studied research problem. Image blur occurs due to relative motion between cameras and objects in the scene during the exposure time, or due to suboptimal focal settings. The blur process can be mathematically represented as
\begin{equation}\label{eq:model}
    y = B(x)+ n,
\end{equation}
where $B$ is the blur operator producing blurry image  $y$,  $x$ refers to the latent  sharp image to be recovered, and  $n$ the additive noise. When the blur is uniform, the blur operation can be characterized using a convolution with blur kernel $b$
\begin{equation}\label{eq:model}
    y = x* b+ n,
\end{equation}
Even for non-blind deblurring i.e with known  blur operator, sharp image recovery is an ill-posed problem. When the blur operator is also unknown, it is referred to as blind deblurring, which is even more severely ill-posed, as multiple pairs
of $B$ and $x$ can produce the same blurred observation $y$. \begin{figure}[ht]
\begin{center}
\resizebox{0.8\linewidth}{!}{
\begin{tabular}{c}
\includegraphics[width=0.6\linewidth]{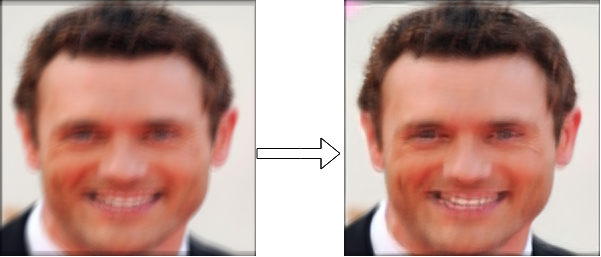}~~
\includegraphics[width=0.26\linewidth]{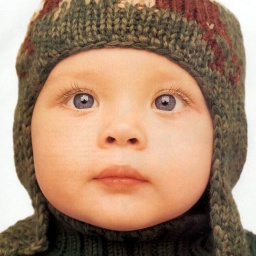}\\
a)~Deblurring without noise~~~~~~~~~~~~~c)~Target image\\
\includegraphics[width=0.6\linewidth]{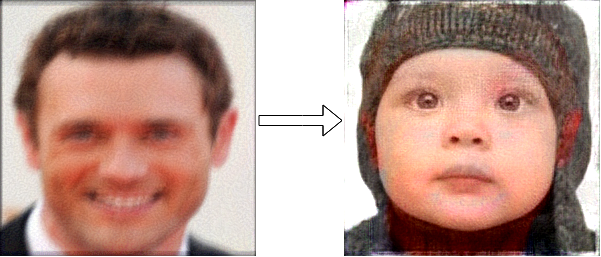}~~\includegraphics[width=0.26\linewidth]{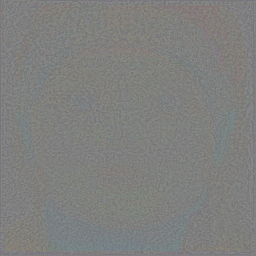}\\
b)~Deblurring with adversarial input~~~~~~d)~Adversarial noise

\end{tabular}}
\caption{Example targetted attack on DeblurGANv2\cite{kupyn2019deblurgan}. 
}
\label{fig:attack_teaser}
\end{center}
\vspace{-0.25em}
\end{figure}
Classical approaches to deblurring employ energy  minimization with suitable priors, e.g. \cite{krishnan2009fast}, and obtain the latent image  $x$ using iterative algorithms.  Blind deblurring methods \cite{jin2018normalized,chen2019blind} employ alternate minimization schemes to recover both $x$ and $B$.  Following the success of deep neural networks for  computer vision applications, recently deep learning approaches have  become the state of the art in  image deblurring and other image restoration tasks \cite{zamir2021multi,kupyn2019deblurgan,gong2020learning,eboli2020end,dong2020deep}. 
While end to end trained deep networks  can achieve impressive performances in many computer vision applications, these networks have been shown to be vulnerable to adversarial examples, wherein addition of carefully crafted imperceptible perturbation to the inputs can produce bizarre results \cite{szegedy2013intriguing,goodfellow2014explaining,aleksandermadry}. Recent works \cite{antun2019instabilities,Choi_2019_ICCV} have demonstrated  adversarial attacks on medical image reconstruction and super-resolution.  

In this work, we introduce adversarial attacks on image deblurring networks. We consider both blind and non-blind deblurring deep networks  and evaluate their robustness to adversarial attacks. While most existing works focus on non-targeted attacks on image reconstruction, we investigate susceptibility of image deblurring networks to both targeted attacks and untargeted attacks. We demonstrate that state of the art deblurring networks are highly susceptible to adversarial perturbations. Fig.~\ref{fig:attack_teaser} shows an example targeted attack on DeblurGANv2~\cite{kupyn2019deblurgan}, a popular image deblurring network. A tiny additive perturbation to the input is sufficient to change the network output from the image of a man to that of a baby.  While non-blind methods are relatively stable to such extreme changes in content, they can be susceptible to localized changes and  untargeted attacks, indicating necessity for robust networks for image recovery.
\section{Related Work}
\label{sec:related work}
\textbf{Deep learning based image deblurring:~} 
Recently image restoration has witnessed a paradigm shift from classical approaches to using deep neural networks. We refer to \cite{KOH2021103134} for a detailed survey and comparison of deep learning based image deblurring methods. Neural network approaches to blind deblurring typically learn to invert the blur operation directly using a trained neural network \cite{kupyn2019deblurgan, zamir2021multi} from large datasets of sharp and blurry image pairs to recover clean images. However, there are also methods which explicitly include the estimation of a blur operator \cite{schuler2015learning,chakrabarti2016neural}. Even for non-blind deblurring, the knowledge of a blur operator has been successfully integrated into neural networks, by unrolling fixed steps of optimization algorithms with learned operators \cite{gong2020learning,eboli2020end,bertocchi2020deep}, or by using known deconvolution techniques in feature space \cite{dong2020deep} within the network. 
Recent works \cite{Vasu_2018_CVPR,nan2020deep} also take into account kernel uncertainty in non-blind deblurring. In addition to end to end trained networks for image recovery, neural networks  are also used in iterative image recovery e.g. by using trained denoisers as proximal operators for plug and play reconstruction \cite{Meinhardt_2017_ICCV}, or using trained generative priors \cite{asim2020blind}.\vspace{2pt}\\
\textbf{Adversarial attacks on image reconstruction:~}
While adversarial attacks on a neural network  have  been first introduced and extensively studied  in the context of image classification \cite{szegedy2013intriguing,goodfellow2014explaining,aleksandermadry}, recent works \cite{antun2019instabilities,pmlr-v119-raj20a} have extended this to image reconstruction. While \cite{antun2019instabilities} investigates instabilities of MRI reconstruction networks by adding perturbations in the image domain, \cite{pmlr-v119-raj20a} consider adversarial perturbations in measurement domain and perform adversarial training using auxiliary network to generate adversarial examples. \cite{pmlr-v121-cheng20a} perform  adversarially attack to generate  tiny features which cannot be recovered well by MRI reconstruction networks and propose adversarial training to improve the network's sensitivity to  such features.  \cite{genzel2020solving}  investigate adversarial robustness of different model based and model-inspired networks for CT and MRI reconstruction to untargeted attacks. They also investigate if networks can be attacked such that the resulting reconstruction causes a misclassification.  In context of image restoration, \cite{Choi_2019_ICCV} shows that several state of the art trained networks for image super resolution are susceptible to adversarial perturbations, however they consider only direct inversion models, with a focus on untargeted attacks. To the best of our knowledge, such attacks have not been shown for deblurring.  
\section{Stability of Image Reconstruction: Adversarial Attacks}
Consider the problem  of reconstructing $x$ from the measurement model \eqref{eq:model}. An ideal reconstruction operator or a network $\net$ should have some notion of Lipschitz continuity such that small changes in the input produce only small bounded changes in the result.
\begin{equation*}
    \|\net(y_1)-\net(y_2)\| \leq K\|y_1-y_2\|
\end{equation*}
Such error estimates exist (in terms of Bregman distances) in the case of convex energy minimization methods, (c.f.~\cite{burger2007error}, Theorem 3.1 \& the formula thereafter). While it is difficult to precisely characterize this notion for deep neural networks, prior works c.f.~\cite{combettes2020lipschitz} attempt to approximate  Lipschitz constant of neural networks. 
Moreover, instabilities could also arise from inaccurate estimate of the measurement operator, which can affect methods incorporating model knowledge into their architecture. 

In this work, we investigate the robustness of image reconstruction networks  to adversarial examples. Specifically, we craft adversarial examples by adding tiny norm bounded perturbations in the measurement domain. We consider only white box attacks where the parameters of the reconstruction network $\net$ are known to the attacker. \vspace{2pt}
\\\textbf{Untargeted Attacks:~}
 For a fixed image $x$, and a reconstruction network $\net$, the goal of an untargeted attack is to find an additive image perturbation that maximizes the reconstruction error subject to $L_\infty$ constraints on the perturbation,
\begin{equation}\label{eq:untargetted}
	\delta_{adv} = \argmax_{\delta\in\mathrm{R}^m} \norm{\net(y + \delta) - \net(y)}_2 \text{  s.t.  }  \norm{\delta}_\infty \leq \epsilon.
\end{equation}
\textbf{Targeted Attacks:}
Here the goal is to find an additive image perturbation that produces a reconstruction close to a target image $\tilde x$ subject to  $L_\infty$ constraints on the perturbation,
\begin{equation}\label{eq:targetted}
	\delta_{adv} = \argmin_{\delta\in\mathrm{R}^m} \norm{\net(y + \delta) - \tilde x }_2 \text{  s.t.  }  \norm{\delta}_\infty \leq \epsilon.
\end{equation}
We solve the constrained optimization problems \eqref{eq:untargetted}, \eqref{eq:targetted} using the projected gradient descent (PGD) algorithm \cite{aleksandermadry}, with gradient updates using the Adam optimizer \cite{kingma2015adam}.
\begin{figure*}[htb]
\begin{center}
\small
\resizebox{\linewidth}{!}{
\begin{tabular}{c}
\hspace{-5pt}\includegraphics[width=0.07\linewidth]{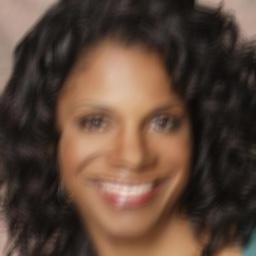}
\includegraphics[width=0.07\linewidth]{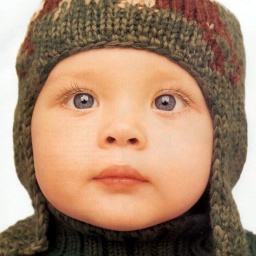}
\includegraphics[width=0.07\linewidth]{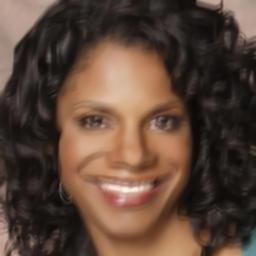}
\includegraphics[width=0.07\linewidth]{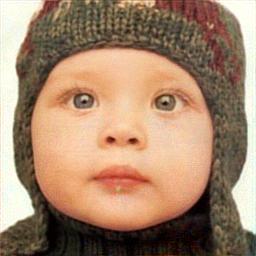}
\includegraphics[width=0.07\linewidth]{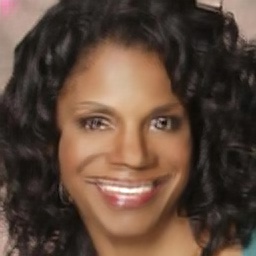}
\includegraphics[width=0.07\linewidth]{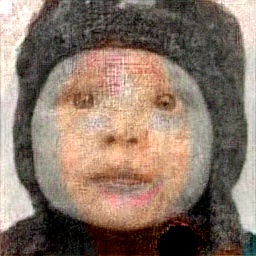}
\includegraphics[width=0.07\linewidth]{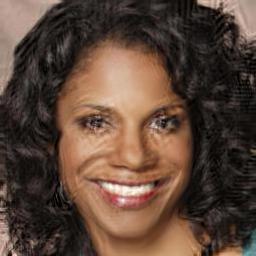}
\includegraphics[width=0.07\linewidth]{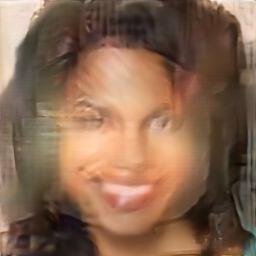}
\includegraphics[width=0.07\linewidth]{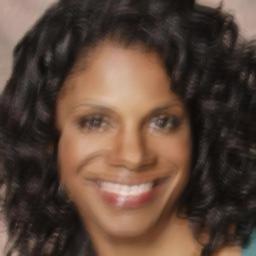}
\includegraphics[width=0.07\linewidth]{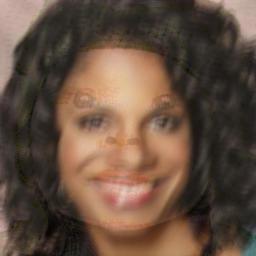}
\\
\hspace{-5pt}\includegraphics[width=0.07\linewidth]{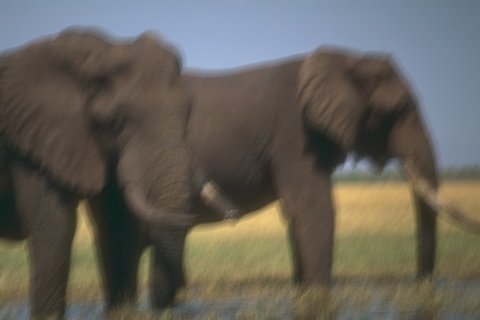}
\includegraphics[width=0.07\linewidth]{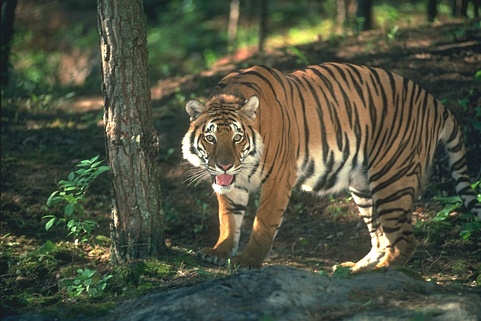}
\includegraphics[width=0.07\linewidth]{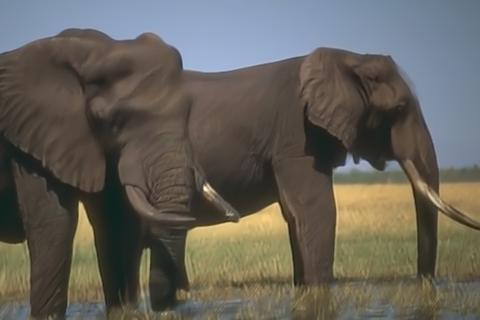}
\includegraphics[width=0.07\linewidth]{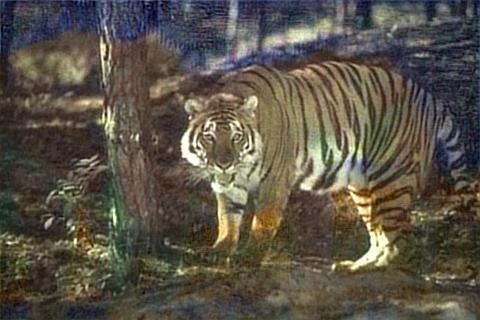}
\includegraphics[width=0.07\linewidth]{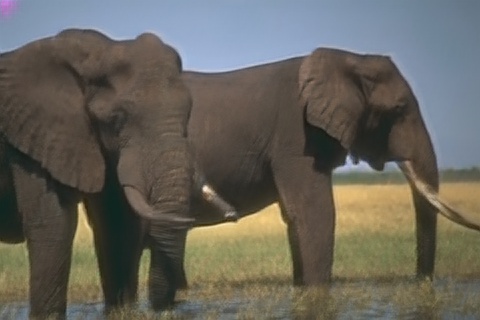}
\includegraphics[width=0.07\linewidth]{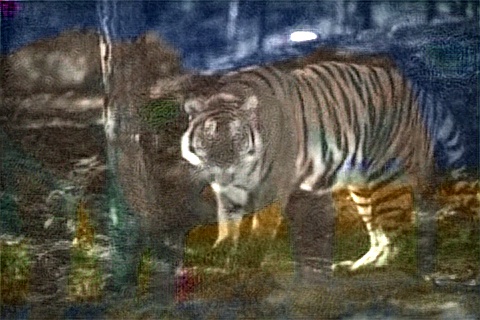}
\includegraphics[width=0.07\linewidth]{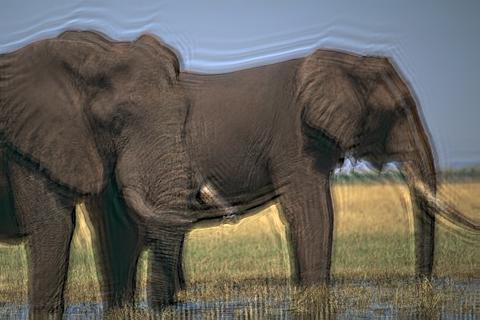}
\includegraphics[width=0.07\linewidth]{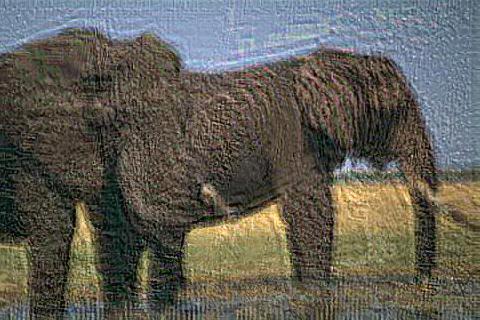}
\includegraphics[width=0.07\linewidth]{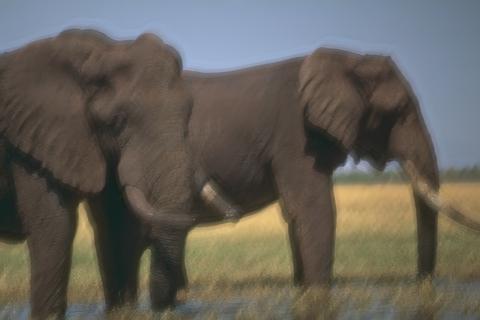}
\includegraphics[width=0.07\linewidth]{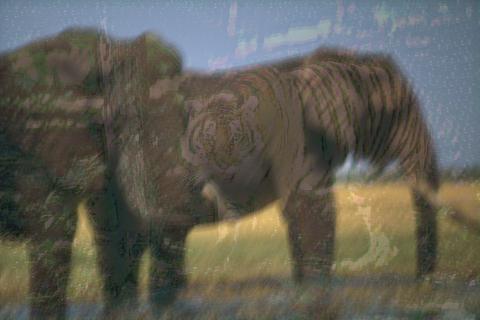}\\
\resizebox{0.7\linewidth}{!}{\begin{tabular}{cccccccccc}
\hspace{-20pt}Blurred~~~~~~~~&Target~&~~~~\cite{zamir2021multi}~Clean&~~~~\cite{zamir2021multi}~$\epsilon=4.$&~~~\cite{kupyn2019deblurgan}~Clean&~~\cite{kupyn2019deblurgan}~$\epsilon=4.$&~~~~\cite{dong2020deep}~Clean&~~\cite{dong2020deep}~$\epsilon=8.$&~~~\cite{gong2020learning}~Clean&~~~\cite{gong2020learning}~$\epsilon=12.$
\end{tabular}}
\end{tabular}}
\caption{Example targeted adversarial attacks on deblurring networks \cite{zamir2021multi,kupyn2019deblurgan,dong2020deep,gong2020learning}.The columns 1 and 2 are blurred inputs and target images. For each network  clean outputs (left), and network outputs with adversarial inputs (right) are depicted. Blurred images in rows 1 and 2 are generated using blur kernels `1' and `2' of size $19\times19$ and $17\times17$ in the dataset of \cite{patchdeblur_iccp2013} respectively.}\label{fig:target_drastic}
\end{center}
\vspace{-0.25em}
\end{figure*}
\begin{figure}[htb]
\begin{center}
\resizebox{\linewidth}{!}{
\begin{tabular}{cccc}
\hspace{-5pt}
\includegraphics[width=0.24\linewidth]{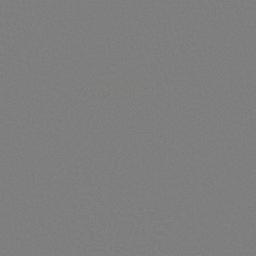}&\hspace{-10pt}
\includegraphics[width=0.24\linewidth]{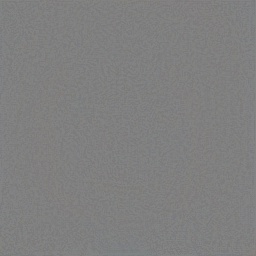}&\hspace{-10pt}
\includegraphics[width=0.24\linewidth]{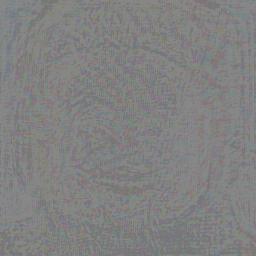}&\hspace{-10pt}
\includegraphics[width=0.24\linewidth]{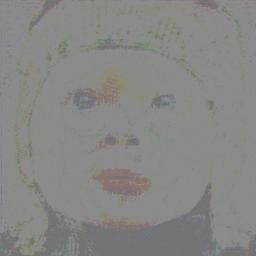}\\
\cite{zamir2021multi}
$\epsilon$=4.&\cite{kupyn2019deblurgan} $\epsilon$=4&\cite{dong2020deep} $\epsilon$=8&\cite{gong2020learning} $\epsilon$=12.\\
\end{tabular}}
\caption{Crafted adversarial perturbation for targeted attacks.}\label{fig:noise_target}
\end{center}
\end{figure}
\begin{figure}[htb]
\begin{center}
\resizebox{\linewidth}{!}{
\begin{tabular}{ccccc}
\hspace{-5pt}
\includegraphics[width=0.2\linewidth]{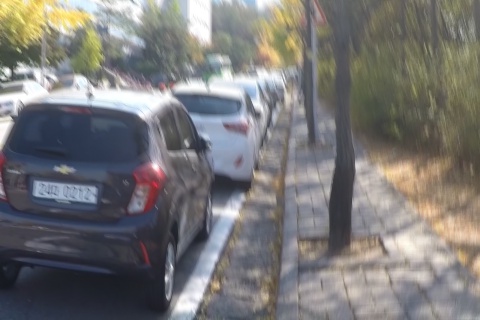}&\hspace{-12pt}
\includegraphics[width=0.2\linewidth]{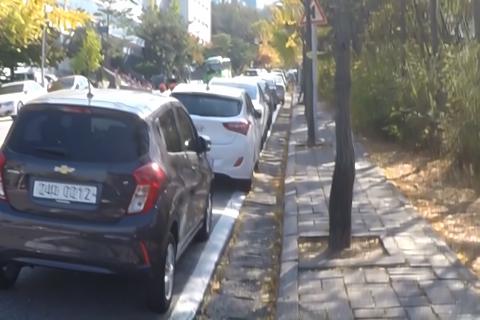}&\hspace{-12pt}
\includegraphics[width=0.2\linewidth]{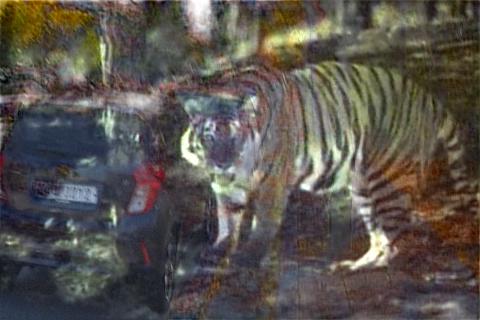}&\hspace{-12pt}
\includegraphics[width=0.2\linewidth]{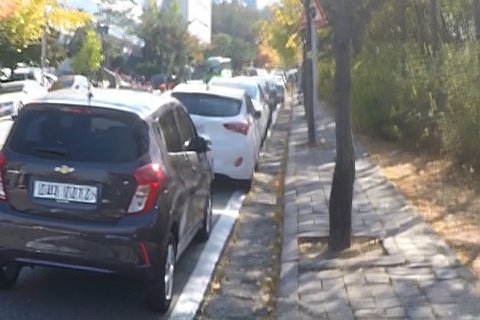}&\hspace{-12pt}
\includegraphics[width=0.2\linewidth]{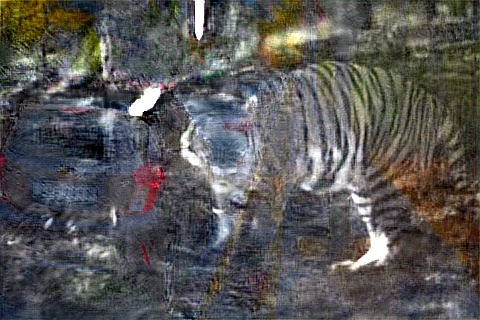}\\
Blurred&
\cite{zamir2021multi} clean& \cite{zamir2021multi} $\epsilon$=4. & \cite{kupyn2019deblurgan} clean &
\cite{kupyn2019deblurgan} $\epsilon$=4.\\
\end{tabular}}
\vspace{-0.5em}
\caption{Targeted attack with dynamically blurred input.}\label{fig:gopro_target}
\vspace{-0.25em}
\end{center}
\end{figure}
\begin{figure}[htb]
\begin{center}
\footnotesize
\resizebox{\linewidth}{!}{
\begin{tabular}{c}
\hspace{-5pt}\begin{sideways}$\epsilon$=4/255\end{sideways}~
\includegraphics[width=0.2\linewidth]{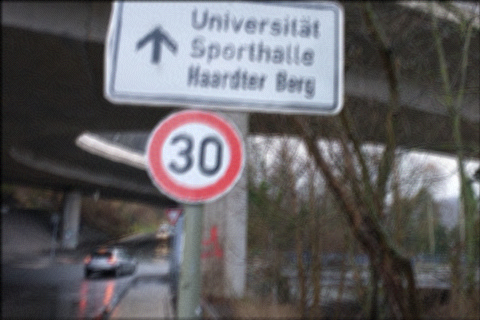}
\includegraphics[width=0.2\linewidth]{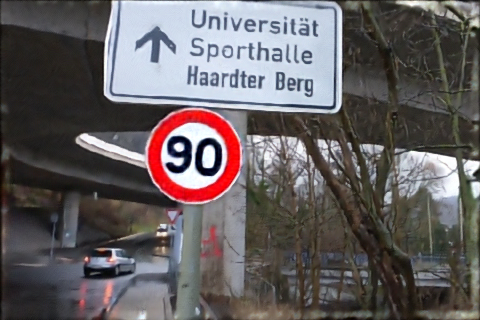}
\includegraphics[width=0.2\linewidth]{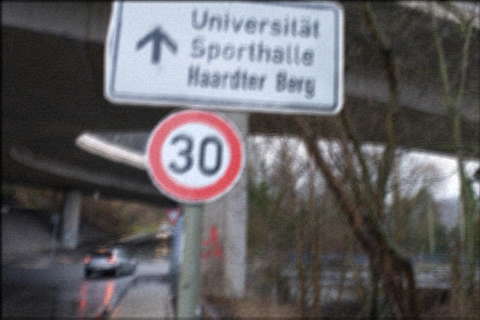}
\includegraphics[width=0.2\linewidth]{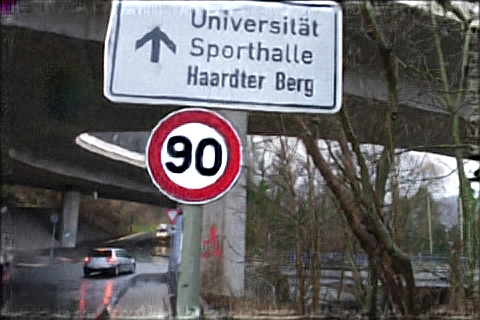}\\
a)~MPR Net\cite{zamir2021multi}~~~~~~~~~~~~~b)~DeblurGAN\cite{kupyn2019deblurgan}\\
\hspace{-5pt}\begin{sideways}$\epsilon$=4/255\end{sideways}~
\includegraphics[width=0.2\linewidth]{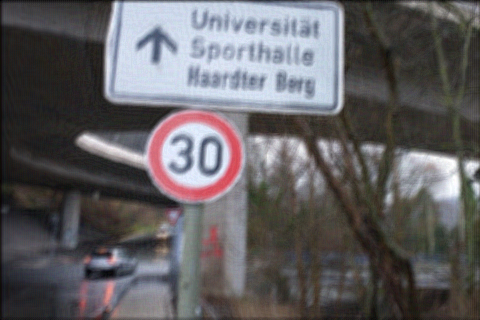}
\includegraphics[width=0.2\linewidth]{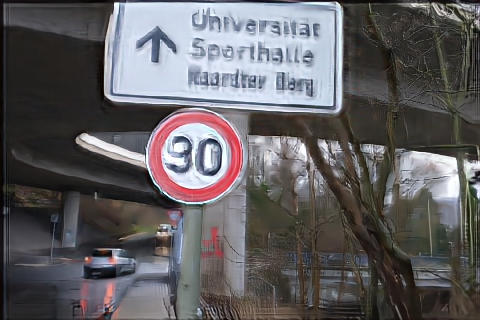}
\includegraphics[width=0.2\linewidth]{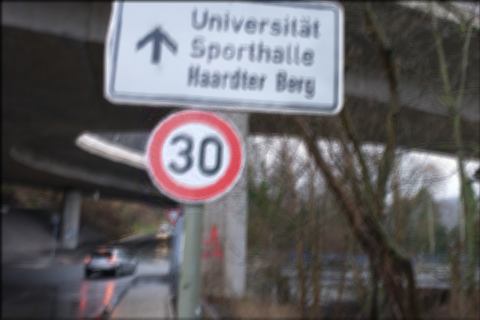}
\includegraphics[width=0.2\linewidth]{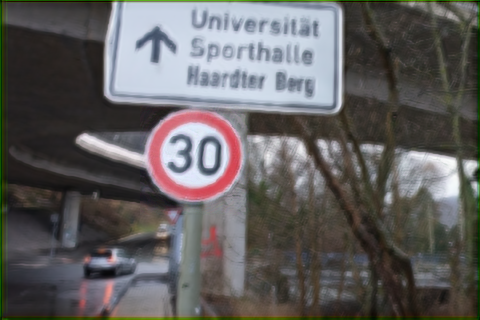}\\
\hspace{-5pt}\begin{sideways}$\epsilon$=8/255\end{sideways}~
\includegraphics[width=0.2\linewidth]{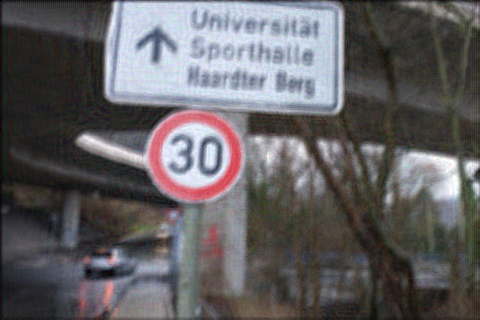}
\includegraphics[width=0.2\linewidth]{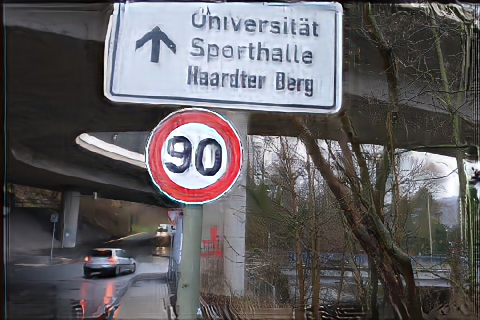}
\includegraphics[width=0.2\linewidth]{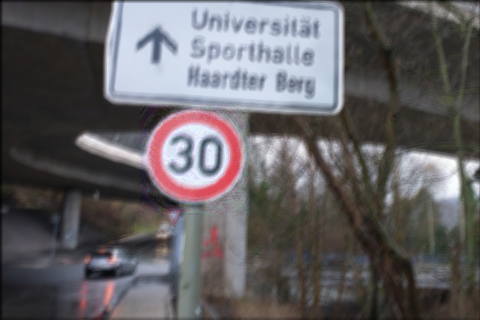}
\includegraphics[width=0.2\linewidth]{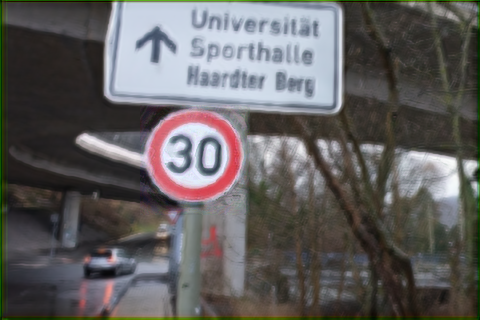}\\
\hspace{-5pt}\begin{sideways}$\epsilon$=12/255\end{sideways}~
\includegraphics[width=0.2\linewidth]{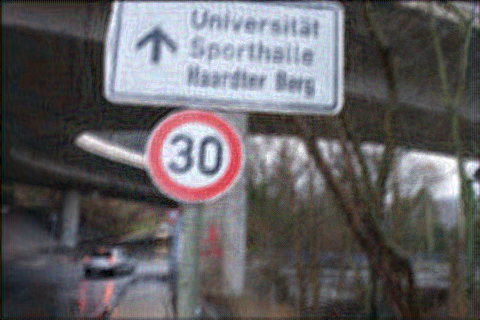}
\includegraphics[width=0.2\linewidth]{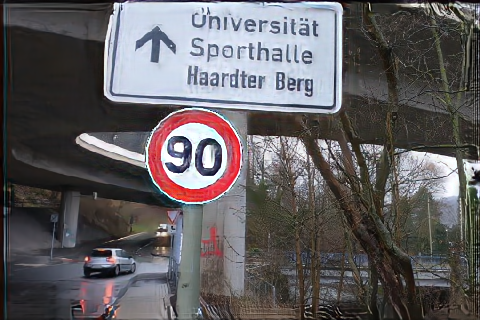}
\includegraphics[width=0.2\linewidth]{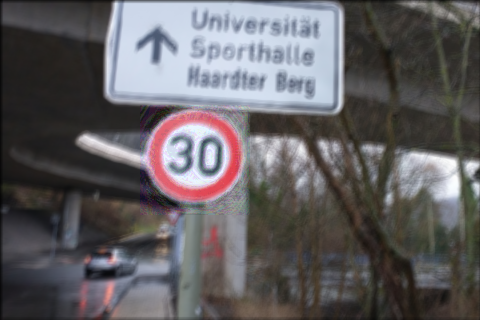}
\includegraphics[width=0.2\linewidth]{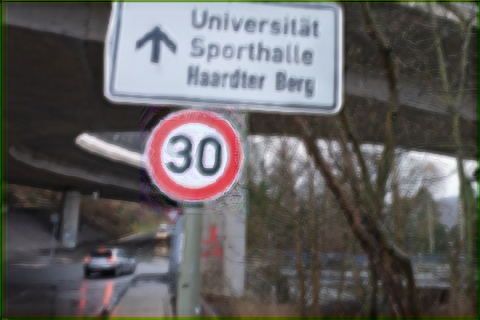}\\
c)~Deep Wiener\cite{dong2020deep}~~~~~~~~~~~~~d)~RGDN\cite{gong2020learning}\\
\end{tabular}}
\caption{Illustration of localized targeted attack. For each approach, the first column is the adversarial input and the second column is the network prediction.}\label{fig:vary_eps}
\end{center}
\vspace{-0.25em}
\end{figure}

\section{Experiments}
We use the following networks in our experiments: i)~DeblurGANv2~\cite{kupyn2019deblurgan} and ii)~MPRNet~\cite{zamir2021multi}, which are end to end trained networks for blind image deblurring, as well as iii)~\cite{gong2020learning}, a learned recurrent gradient descent network, and iv) \cite{dong2020deep}, which performs Wiener deconvolution in the feature space of neural networks. The non-blind deblurring networks \cite{gong2020learning,dong2020deep} use the knowledge of blur operator during reconstruction. We use publicly available trained models of all these networks made available by the authors. For attacks on DeblurGANv2~\cite{kupyn2019deblurgan}, we choose the version using the inception backbone since it achieves the  best results.
In their experiments, \cite{gong2020learning} can unroll the recurrent gradient descent network for arbitrary number of steps during test time till a stopping criteria is satisfied. However, it becomes prohibitively complex to perform adversarial attack for high number of unrolled steps. In our experiments, we limit the number of unrolled steps to 10 for crafting adversarial inputs, but evaluate robustness to the same inputs using a network with 50 unrolled steps. We create a synthetic dataset of blurred images generated by convolving 8 uniform blur kernels of dataset in ~\cite{patchdeblur_iccp2013} with  sharp images  of different classes, $5$ images CelebA-HQ dataset resized to $256\times256$, 
 and 5 images from Berkeley segmentation dataset (BSDS300). These blur kernels have sizes ranging from $13\times13$ to $27\times27$. 
In our experiments, we use a fixed step size of $1e-3$ and use 250 PGD steps  and 500 PGD steps  to craft untargeted and targeted adversarial perturbations.
\subsection{Targeted Attacks}
In this section, we investigate the robustness of deblurring networks to targeted attacks that try to make the networks generate an image that is close to a target image. Fig.~\ref{fig:target_drastic} illustrates example targeted attacks on the deblurring methods \cite{zamir2021multi,kupyn2019deblurgan,dong2020deep,gong2020learning}. Even though the blind deblurring methods \cite{zamir2021multi,kupyn2019deblurgan} are not trained using uniform blur models, they generate sharper images with less visible artifacts when the inputs are clean and the blur kernels are not too large. However, they are also most susceptible to targeted attacks, shockingly turning a woman into a baby or an elephant into a tiger with adversarial noise strength as low as 4/255. 

In contrast, the non-blind methods are more robust and do not produce such extreme changes in the output, even for higher strengths of adversarial noise. The additive adversarial noise causing the targeted attacks is illustrated in Fig.~\ref{fig:noise_target}. Adversarial perturbation for the non-blind methods show clear pattern of source images.  However, the visual quality of the non-blind network outputs \cite{dong2020deep,gong2020learning} is lower even without adversarial noise. On our test data, the deep Wiener network \cite{dong2020deep} produces sharper results, but with visible ringing artifacts, and the recurrent gradient decent network \cite{gong2020learning} outputs still have a residual blur effect. 
Quantitative evaluation provided in Tab.~\ref{tab:targetted} confirms the  trend of higher susceptibility of the blind deblurring approaches  to targeted attacks, showing higher similarity with the target image in terms of PSNR and normalized cross correlation (NCC) than with respect to the actual ground truth. While the blind dynamic deblurring approaches \cite{zamir2021multi,kupyn2019deblurgan} are not trained using uniform blur models, we find that similar adversarial vulnerabilities occur even with dynamically blurred images from the test sets of  \cite{zamir2021multi,kupyn2019deblurgan}, see Fig.~\ref{fig:gopro_target}.

\begin{table*}[htb]
    \centering
    \resizebox{\linewidth}{!}{
\begin{tabular}{|c|c|c|ccc|ccc||ccc|}
\hline
\hspace{-5pt}\multirow{3}{*}{Data}&\multirow{3}{*}{Net}&\multirow{3}{*}{Clean}&\multicolumn{6}{c||}{Targeted attacks}&\multicolumn{3}{c|}{Untargeted attacks}\\
\cline{4-12}
&&&
\multicolumn{3}{c|}{Similarity to source PSNR/NCC}&\multicolumn{3}{c||}{Similarity to target PSNR/NCC} &\multicolumn{3}{c|}{Similarity to source PSNR/NCC}\\
     &&&$\epsilon=4$&$\epsilon=8$&$\epsilon=12$&$\epsilon=4$&$\epsilon=8$&$\epsilon=12$&$\epsilon=4$&$\epsilon=8$&$\epsilon=12$\\
    \hline
    \hspace{-5pt}\multirow{4}{*}{Faces}
    &\cite{zamir2021multi}&26.81/0.976&
9.77/0.409&9.75/ 0.408&9.75/0.408&26.80/ 0.986&26.94/0.987&26.94/0.987& 11.29/ 0.645&8.88/0.498&8.582/0.465\\
&\cite{kupyn2019deblurgan}&27.13/0.982& 10.26/0.419&10.09/0.413&10.081/0.412& 20.57/0.950&
21.66/0.963& 21.76/0.963& 5.65/-0.063& 5.36/-0.137&5.23/-0.175\\
&\cite{dong2020deep}&22.91/0.951&19.28/0.906&17.07/0.858&15.58/0.810&11.31/0.565&12.55/0.659&13.69/0.728&11.65/0.593& 10.12/0.511& 9.10/0.456\\
&\cite{gong2020learning}&26.98/0.981&24.55/0.961&23.32/0.954&22.06/0.934&10.19/0.452& 10.57/0.496&10.95/0.537&24.13/0.966&22.67/0.953&21.43/0.939\\
    \hline
        \hspace{-5pt}\multirow{4}{*}{BSD}
   &\cite{zamir2021multi}&24.41/ 0.944&12.52/0.155&12.30/0.130&12.27/0.126&23.15/0.899&24.14/0.919&24.22/0.921&10.59/0.487&9.04/0.398&8.12/0.319\\
&\cite{kupyn2019deblurgan}&24.04/0.941&
12.43/0.174&12.35/0.164& 12.27/0.153&19.94/0.805&20.92/0.841&21.27/0.854&5.90/0.173&5.64/0.147&5.57/0.085\\
&\cite{dong2020deep}&21.44/0.443&21.08/0.882&19.43/0.844&18.08/0.792&13.11/0.139&14.10/0.223&15.00/0.311& 14.15/0.607&11.43/0.478&10.15/0.416\\
&\cite{gong2020learning}&24.23/0.943&22.61/0.921&21.78/0.907&20.89/0.887& 12.83/ 0.106&13.29/0.166&13.75/0.226&22.08/0.907&21.11/0.884&20.22/0.859
\\
\hline
    \end{tabular}}
    \caption{Comparison of  PSNR  and normalized cross correlation (NCC) values with respect to source image for untargeted attacks, PSNR and NCC with respect to source and target images for targeted attacks to evaluate robustness.}
    \label{tab:targetted}
\end{table*}

\begin{figure*}[htb]
\begin{center}
\footnotesize
\resizebox{\linewidth}{!}{
\begin{tabular}{ccccccccc}
\hspace{-5pt}\includegraphics[width=0.1\linewidth]{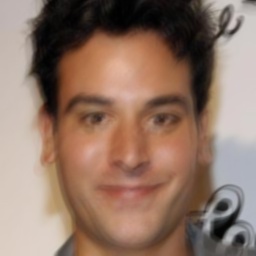}&
\hspace{-10pt}\includegraphics[width=0.1\linewidth]{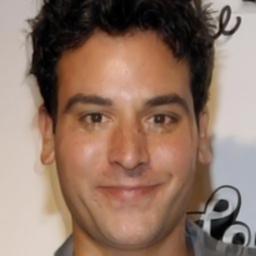}&
\hspace{-10pt}\includegraphics[width=0.1\linewidth]{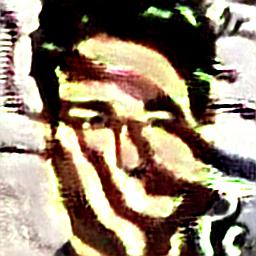}&
\hspace{-10pt}\includegraphics[width=0.1\linewidth]{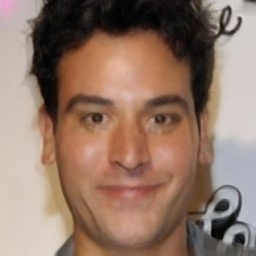}&
\hspace{-10pt}\includegraphics[width=0.1\linewidth]{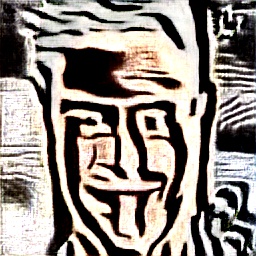}&
\hspace{-10pt}\includegraphics[width=0.1\linewidth]{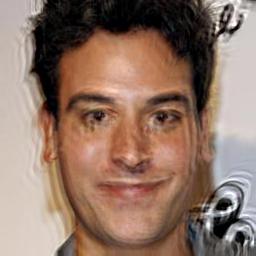}&
\hspace{-10pt}\includegraphics[width=0.1\linewidth]{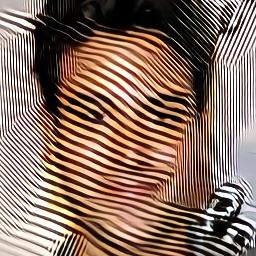}&
\hspace{-10pt}\includegraphics[width=0.1\linewidth]{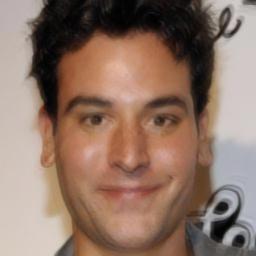}&
\hspace{-10pt}\includegraphics[width=0.1\linewidth]{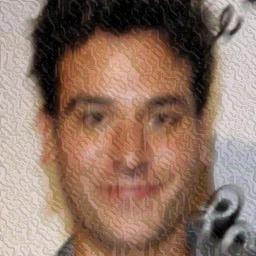}
\\
Blurred&\cite{zamir2021multi}~Clean&\cite{zamir2021multi}~$\epsilon=4.$&\cite{kupyn2019deblurgan}~Clean&\cite{kupyn2019deblurgan}~$\epsilon=4.$&\cite{dong2020deep}~Clean&\cite{dong2020deep}~$\epsilon=4.$&\cite{gong2020learning}~Clean&\cite{gong2020learning}$\epsilon=8.$
\end{tabular}}
\caption{Untargeted adversarial attack on deblurring networks. Blur kernel `6' of size $21\times21$ in the dataset of \cite{patchdeblur_iccp2013} is used.}\label{fig:untargeted}
\end{center}
\vspace{-0.25em}
\end{figure*}
\begin{figure}[htb]
\begin{center}
\small
\resizebox{0.8\linewidth}{!}{
\begin{tabular}{c}
\hspace{-10pt}\begin{sideways}~~Clean obs.\end{sideways}~\includegraphics[width=0.28\linewidth]{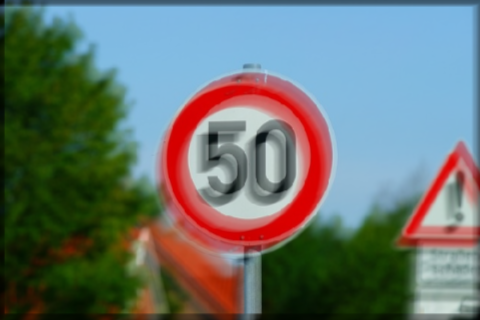}
\includegraphics[width=0.28\linewidth]{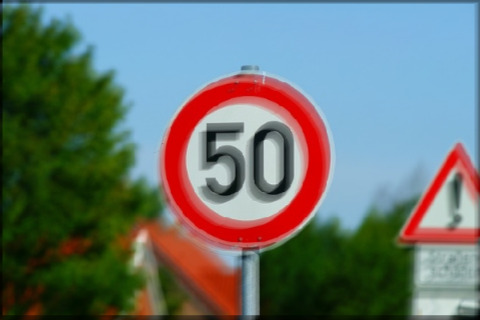}
\includegraphics[width=0.28\linewidth]{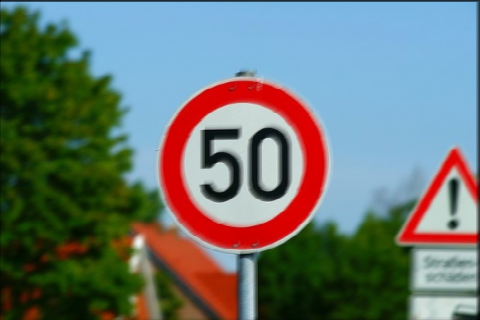}\\
\hspace{-10pt}\begin{sideways}~~~~~~~~\cite{dong2020deep}\end{sideways}~\includegraphics[width=0.28\linewidth]{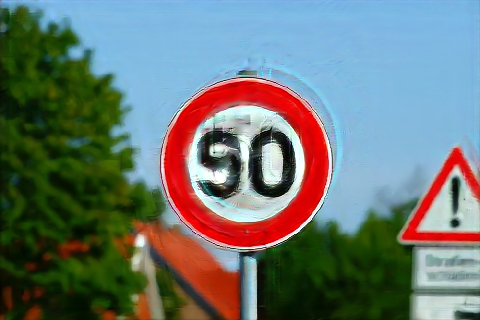}
\includegraphics[width=0.28\linewidth]{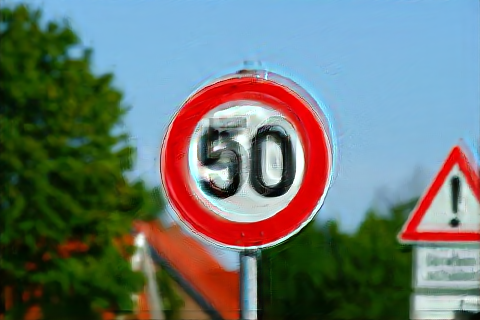}
\includegraphics[width=0.28\linewidth]{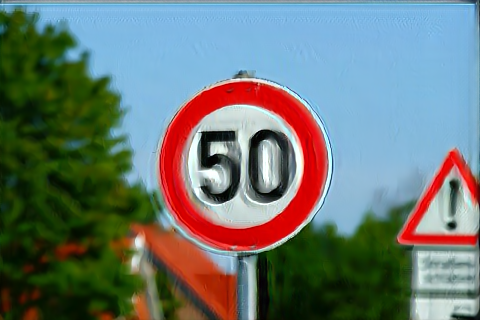}\\

\hspace{-10pt}\begin{sideways}~~~~~~~~\cite{gong2020learning}\end{sideways}~\includegraphics[width=0.28\linewidth]{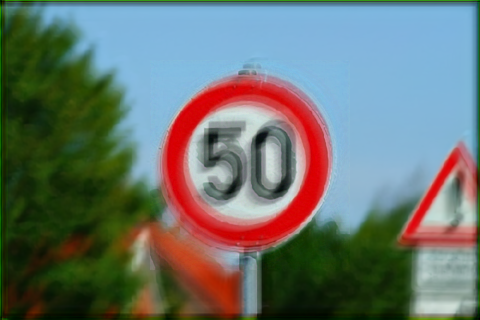}
\includegraphics[width=0.28\linewidth]{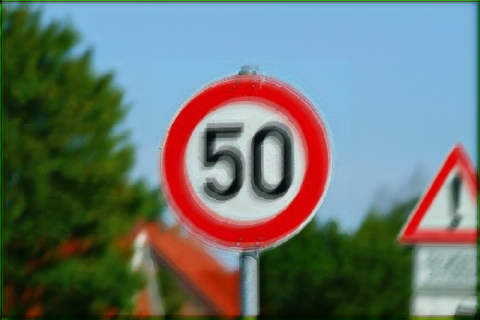}
\includegraphics[width=0.28\linewidth]{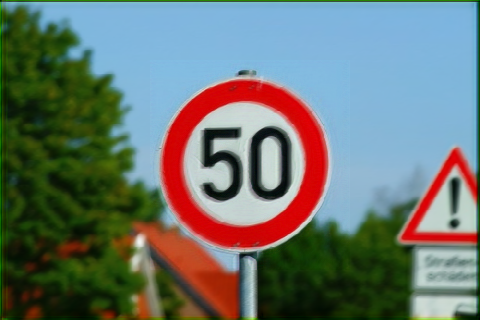}\\
$k=25$~~~~~~~~~~~~~~~~~~~$k=17$~~~~~~~~~~~~~~~~$k=11$\\
\end{tabular}}
\caption{Effect of kernel size on ease of targetted attack on non-blind deblurring networks with $\epsilon=4/255$.}\label{fig:vary_kernelsize}
\end{center}
\vspace{-0.25em}
\end{figure}
We now investigate the susceptibility of networks to targeted attacks where target image is modified at a small localized region.
Fig.~\ref{fig:vary_eps} shows such a targeted attack on the deblurring networks, where the target image has the speed limit sign modified from `30' to `90'. The attack on blind deblurring networks is successful even at $\epsilon=4/255$.
Among the non-blind networks, attack on deep Wiener network \cite{dong2020deep} is clearly successful at $\epsilon=12/255$, while the target features begin to manifest at even lower values of $\epsilon$. The learned gradient descent approach \cite{gong2020learning} is most difficult to attack, barely producing target features even for $\epsilon=12/255$. 
As the size of blur kernel becomes larger, deblurring becomes more ill-posed, which can influence the stability of the reconstruction. We investigate this effect by evaluating adversarial robustness of non-blind networks to targeted attacks by fixing adversarial noise level to 4/255, and varying the blur kernel size $\{25,17,11\}$. Here the target has a localized change in the speed limit sign from `50' to `90'. As the blur effect in the input reduces, we expect the networks to be more robust to attacks, which is confirmed by the results in Fig.~\ref{fig:vary_kernelsize}.  The robustness of  deep Wiener deblurring \cite{dong2020deep} improves as the inputs are less and less blurred, and the learned gradient is least susceptible to attack. 
\subsection{Untargeted Attacks}
In Tab.~\ref{tab:targetted} and Fig.~\ref{fig:untargeted} we provide of effect of untargetted attacks on deblurring networks which increase the reconstruction loss.  While the blind deblurring networks are highly susceptible to untargeted attacks, we find even the non-blind method of deep Wiener filtering also being unstable, even at low adversarial noise strengths. 

In all our experiments, we find that the blind deblurring methods \cite{kupyn2019deblurgan,zamir2021multi} are most susceptible to adversarial perturbations. One reason is that blind deconvolution is inherently more ill-posed, making the reconstruction problem more unstable. Moreover, both the methods \cite{kupyn2019deblurgan,zamir2021multi}  use only clean data during training, whereas the methods  \cite{gong2020learning,dong2020deep} also add noise to blurry inputs during training. Recent work  \cite{genzel2020solving} shows addition of noise during training as an effective way to  improve adversarial robustness of deep CT reconstruction. However, training with noise is not sufficient to guarantee adversarial robustness, as seen from the results of deep Wiener deconvolution \cite{dong2020deep}, which is more prone to attacks than the learned gradient descent approach \cite{gong2020learning}. 
\section{Conclusions}
In this paper we introduced adversarial attacks on image deblurring networks and showed that state of the art deblurring methods can be highly susceptible to adversarial attacks. While the performance on clean data is  important, it is critical to improve adversarial robustness of restoration approaches, by robust training or by developing more robust architectures.
\bibliographystyle{IEEEbib}

\end{document}